%% file: eccvw18.tex
\documentclass[runningheads]{llncs}

\usepackage{times}
\usepackage{xcolor}
\usepackage{soul}
\usepackage[utf8]{inputenc}
\usepackage[small]{caption}

\usepackage{epsfig}
\usepackage{graphicx}
\usepackage{amsmath}
\usepackage{amssymb}

\usepackage{adjustbox}
\usepackage{booktabs}
\usepackage{multirow}
\usepackage{array}
\usepackage{caption}

\usepackage{color}
\usepackage[width=122mm,left=12mm,paperwidth=146mm,height=193mm,top=12mm,paperheight=217mm]{geometry}
\begin{document}


\title{Forecasting Hands and Objects in Future Frames}

\titlerunning{Forecasting Hands and Objects in Future Frames}



\author{Chenyou Fan \and Jangwon Lee \and Michael S. Ryoo}
\authorrunning{Chenyou Fan \and Jangwon Lee \and Michael S. Ryoo}

\institute{Indiana University, Bloomington IN 47401, USA 
\email{\{fan6,leejang,mryoo\}@indiana.edu}
}

\maketitle

\begin{abstract}
   This paper presents an approach to \emph{forecast} future presence and location of human hands and objects. Given an image frame, the goal is to predict what objects will appear in the future frame (e.g., 5 seconds later) and where they will be located at, even when they are not visible in the current frame. The key idea is that (1) an intermediate representation of a convolutional object recognition model abstracts scene information in its frame and that (2) we can predict (i.e., regress) such representations corresponding to the future frames based on that of the current frame. We present a new two-stream fully convolutional neural network (CNN) architecture designed for forecasting future objects given a video. The experiments confirm that our approach allows reliable estimation of future objects in videos, obtaining much higher accuracy compared to the state-of-the-art future object presence forecast method on public datasets.
   
\end{abstract}


\input{intro}
\input{related_work}
\input{approach}
\input{experiments}
\input{conclusion}

\bibliographystyle{splncs04}
\bibliography{egbib}

\end{document}

%% file: intro.tex
\section{Introduction}

The ability to forecast future scene is very important for intelligent agents. The idea is to provide them an ability to infer future objects, similar to humans predicting how the objects in front of them will move and what objects would newly appear.
This is particularly necessary for interactive/collaborative systems, including robots, autonomous cars, surveillance systems, and wearable devices. For instance, a robot working on a collaborative task with a human needs to predict what objects the human is expect to move and how they will move; a surgery robot needs to forecast what surgical instruments a human surgeon will need in a few seconds to better support the person. Similarly, forecasting will enable an autonomous driving agent to predict when and where pedestrians or other vehicles are likely to appear even before they are within the view. The forecasting is also necessary for more natural human-robot interaction as well as better real-time surveillance, since this will allow faster reaction of such systems in response to humans and objects.



In the past 2-3 years, there has been an increasing number of works on `forecasting' in computer vision. Researchers studied forecasting trajectories \cite{kitani12,walker2014patch}, convolutional neural network (CNN) representations \cite{vondrick2015anticipating}, optical flows and human body parts \cite{luo2017unsupervised}, and video frames \cite{lotter2016deep,finn2016unsupervised}. However, none of these approaches were optimized for forecasting explicit locations of objects appearing in videos. Vondrick et al.~\cite{vondrick2015anticipating} only forecast presence of objects without giving their future locations. The method of Luo et al.~\cite{luo2017unsupervised} requires the person to be already in the scene in order to forecast his/her future pose, also it does not consider objects. 
\cite{lotter2016deep} and \cite{finn2016unsupervised} were designed for forecasting direct image frames, instead of doing object-level estimations of locations. To our knowledge, an end-to-end approach to learn the forecast model optimized for future object location estimation has been lacking.



\begin{figure*}[t]
	\centering
	\resizebox{0.99\linewidth}{!}{
	  \includegraphics{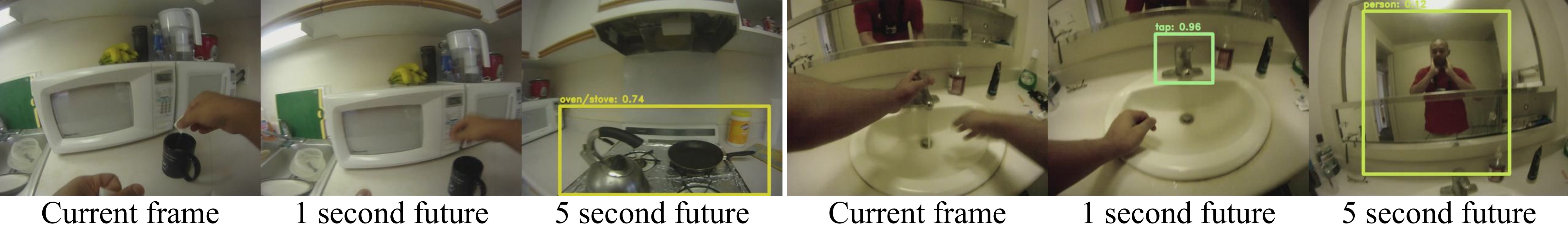}
	}
	\vspace{-3pt}
	\caption{Example object location forecasts on the ADL dataset. Both 1 second and 5 second future bounding box forecasts are presented, overlaid on the actual future frames. The approach only uses the current frame as an input to forecast future boxes. Three types of objects, oven, tap, and person, are predicted in the two examples.}
	\label{fig:adl_examples}
\end{figure*}

This paper introduces a new approach to forecast presence and location of hands and objects in future frames (e.g., 5 seconds later). Given an image frame, the objective is to predict future bounding boxes of appearing objects even when they are not visible in the current frame (e.g., Figure \ref{fig:adl_examples}). Our key idea is that (1) an intermediate CNN representation of an object recognition model abstracts scene/motion information in its frame and that (2) we can model how such representation changes in the future frames based on the training data. For (1), we design a new two-stream CNN architecture with an auto-encoder by extending the state-of-the-art convolutional object detection network (SSD \cite{ssd}). For (2), we present a fully convolutional regression network that allows us to infer future CNN representations. These two networks are combined to directly predict future locations of human hands and objects, forming a deeper network that could be trained in an end-to-end fashion (Figure \ref{fig:our_network}).

We evaluate our proposed approach with two first-person video datasets and one urban street scene dataset. Hands and objects dynamically appear and disappear in first-person videos taken with wearable cameras, making them suitable for the evaluation of our approach. Notably, in our experiments with the public ADL dataset \cite{pirsiavash2012detecting}, our accuracy was higher than the previous state-of-the-art \cite{vondrick2015anticipating} by more than 0.25 mean average precision (AP).


%% file: related_work.tex
\section{Related work}
Researchers are increasingly focusing on `forecasting' of future scene. This work includes early recognition of ongoing human activities \cite{ryoo11,hoai2012max} as well as more explicit forecasting of human trajectories and future locations \cite{kitani12,walker2014patch,ma2016trajectories,park2016future,yagi2017future}. There are also works forecasting future features or video frames themselves \cite{vondrick2015anticipating,finn2016unsupervised,luo2017unsupervised}.

Kitani et al.~\cite{kitani12} presented an approach to predict human trajectories in surveillance videos. \cite{walker2014patch} and \cite{ma2016trajectories} also focused on forecasting trajectories. However, most of these trajectory-based works are limited in the sense that they assume the person/object to be forecasted is already present in the scene. This is particularly limited when dealing with objects recorded in wearable/robot cameras, since objects often go out of the scene and return as the camera and the body parts move. Park et al.~\cite{park2016future} tried to predict the future location of the person, while using egocentric videos. Rhinehart and Kitani~\cite{rhinehart2017iccv} used egocentric videos to learn a reinforcement learning model of the tasks, in order to forecast semantic states and goals. However, their forecasts were done without explicit modeling of when/where the objects need to appear.

Recently, Vondrick et al.~\cite{vondrick2015anticipating} showed that forecasting fully connected layer outputs of a convolutional neural network (e.g., VGG \cite{vgg}) is possible. The paper further demonstrated that such representation forecast can be used to predict the future presence of objects (i.e., whether a particular object will appear within 5 seconds later or not). However, due to the limited dimensionality of the representation (i.e., 4K-D), the approach was not directly applicable for predicting `locations' of objects in the scene. Similarly, Luo et al.~\cite{luo2017unsupervised} used CNN regression to forecast optical flow fields, and used such optical flows to predict how human body part will move. It requires the human to be present in the scene initially and his/her body pose is correctly estimated. Finn et al.~\cite{finn2016unsupervised} predicted future video frames by learning dynamics from training videos, but it also assumed the objects to be already present in the scene.



We believe this is the first paper to present a method to explicitly forecast location of objects in future frames using a fully convolutional network. The contribution of this paper is in (1) introducing the concept of future object forecast using fully convolutional regression of intermediate CNN representations, and (2) the design of the two-stream SSD model to consider both appearance and motion optimized for video-based future forecasting. There were previous works on pixel-level forecasting of future frames including \cite{luo2017unsupervised,finn2016unsupervised,visualdynamics16,walker2016uncertain}, but they were limited to pixel-level motion prediction instead of doing object-level predictions. Our approach does not assume hand/object to be in the scene for their future location prediction, unlike prior works based on tracking (e.g., trajectory-based estimation) or pixel motion (e.g., optical flow estimation). For example, Figure \ref{fig:adl_examples} shows our model forecasting an oven to appear 5-sec later, which is not visible in the current frame.


%% file: approach.tex
\section{Approach}

\begin{figure*}[t]
 \centering
    \includegraphics[width=0.8\textwidth]{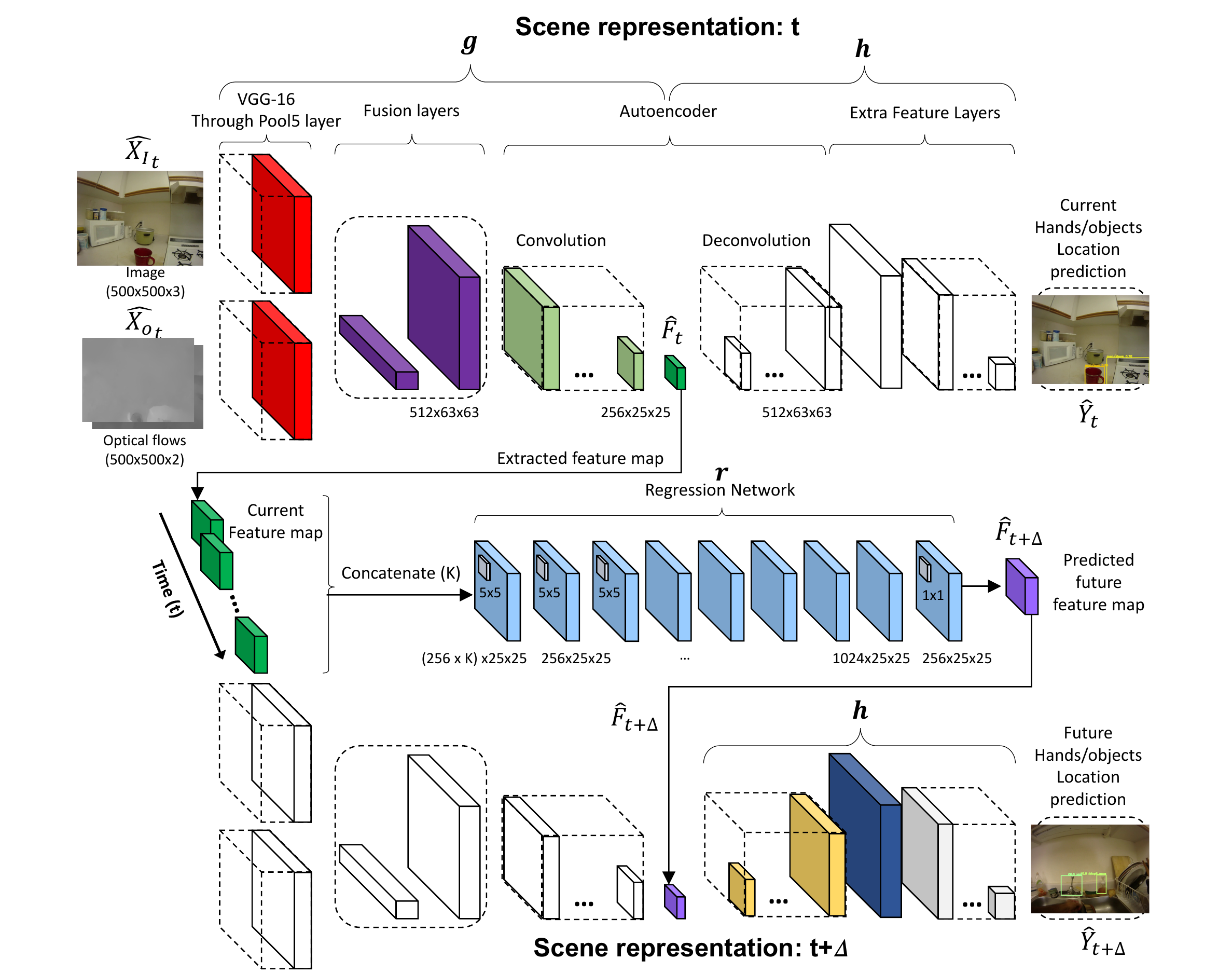}
 \caption{
 Overview of our approach:
 It consists of two fully convolutional neural networks:
 The first network is the two-stream object detection network (the 1st row and the 3rd row of the figure). The 1st row and the 3rd row are the duplicates of the same model.
 The second network is the fully convolutional regression network to predict the future intermediate scene representation (the 2nd row). Only the colored layers are used in the actual testing/inference stage.
 }
 \label{fig:our_network}
\end{figure*}
          
The objective of our approach is to predict hands and objects in the future scene given the current image frame.
We propose a new two-stream convolutional neural network architecture, with a fully convolutional future representation regression module (Figure \ref{fig:our_network}). The proposed model consists of two CNNs:
(1) an extended two-stream video version of the Single Shot MultiBox Detector (SSD) \cite{ssd} also with a convolutional auto-encoder,
and (2) a future regression network to predict the intermediate scene representation corresponding to the future frame.



The key idea of our approach is that we can forecast scene configurations of the near future (e.g., 5 seconds later) by predicting (i.e., regressing) its intermediate CNN representation. Inside our fully convolutional hand/object detection network, we abstract scene/motion information of the input frame as its intermediate representation (i.e., $\mathbf{\hat{F}_t}$ in Figure \ref{fig:our_network}) using convolutional auto-encoder. Our approach estimates the intermediate representation of the `future frame' (i.e., $\mathbf{\hat{F}}_{t+\Delta}$) from it, and combines it with the later layers of the object detection network to obtain their future bounding boxes.

\subsection{Two-stream network for scene representation}

In this subsection, we introduce our two-stream CNN extending the previous fully convolutional object detection network. The objective of this component is to abstract the scene at time $t$ into a lower dimensional representation, so that estimation of hand and object locations becomes possible.



Our two-stream CNN is designed to combine evidence from both spatial and motion information to represent the scene, as shown in the top row of Figure \ref{fig:our_network}. The spatial stream receives the RGB input, while the temporal stream receives the corresponding X and Y gradients of optical flows. This design was inspired by the two-stream network of Simonyan and Zisserman~\cite{simonyan2014two}, which was originally proposed for activity recognition. The intuition behind the use of the two-stream network is that it allows capturing of temporal motion patterns in activity videos as well as spatial information. We used OpenCV TVL1 optical flow algorithm to extract flow images. 



We extend the SSD object detection network for our forecast task. We first insert a fully convolutional auto-encoder to our model, including five convolutional layers followed by five deconvolutional layers. We extract feature maps from the bottleneck layer as our compact scene representation. Each convolutional and deconvolutional layers of the auto-encoder has 5 $\times$ 5 learnable filters. The number of filters in the convolutional layers are: 512, 256, 128, 64, and 256. The deconvolutional layers have the symmetric number of filters: 256, 64, 128, 256, and 512. No pooling is applied and instead, downsampling is achieved by convolution operation with stride=2. This design allows the abstraction of scene information in an image frame into a lower dimensional (256x25x25) intermediate representation.

We design our object-based scene representation network to have both the spatial-steam and temporal-stream part. Instead of using a late-fusion to combine spatial and temporal streams at the end of network as was done in \cite{simonyan2014two}, we design an early-fusion by combining two streams' feature maps before the encoder-decoder component. Specifically, at the conv5 layers, two $256\times25\times25$ feature blobs from both streams are combined to form a single $256\times25\times25$ blob by learning one-by-one kernels. 

In addition, since our future regression component (to be described in Subsection \ref{subsec:regression}) handles combining representations of multiple past frame, we reduce the amount of computations in our temporal stream by making it receive only one optical flow image instead of stacked optical flows from multiple frames.


Let $f$ denote the proposed two-stream CNN to estimate object locations given a video frame at time $t$.
This function has two input variables $\mathbf{\hat{X_I}}_{t}$ and $\mathbf{\hat{X_O}}_{t}$,
which represent the given image frame and the corresponding optical flow image at time $t$ respectively. Note that $\mathbf{\hat{X_O}}_{t}$ is calculated from image $I_{t-1}$ and $I_t$, so no future information after time $t$ is used.
Then, we can decompose this function into two sub functions, $f = g \circ h$:
\begin{equation} \label{eq:1}
\mathbf{\hat{Y}}_{t} = f(\mathbf{\hat{X_I}}_{t}, \mathbf{\hat{X_O}}_{t}) = h(\mathbf{\hat{F}}_{t}) = h(g(\mathbf{\hat{X_I}}_{t}, \mathbf{\hat{X_O}}_{t})),
\end{equation}
where a function $g:(\mathbf{\hat{X_I}}, \mathbf{\hat{X_O}}) \rightarrow \mathbf{\hat{F}}$
denotes convolutional layers to extract compressed visual representation (feature map) $\mathbf{\hat{F}}$
from $\mathbf{\hat{X_I}}_{t}$ and $\mathbf{\hat{X_O}}_{t}$,
and $h:\mathbf{\hat{F}} \rightarrow \mathbf{\hat{Y}}$
indicates the remaining part of the proposed network that uses the compressed feature map as an input for predicting hands and object locations $\mathbf{\hat{Y}}_{t}$ at time $t$. The first row and the last row of Figure \ref{fig:our_network} shows such architecture. 

The loss function used for the training is identical to the original SSD \cite{ssd}, which is a combination of localization and confidence losses. 

\subsection{Future regression network}
\label{subsec:regression}


Our objective is to forecast the locations of objects in the future frame $\mathbf{\hat{Y}}_{t+\Delta}$ based on current frame $\mathbf{\hat{Y}}_{t}$. 
We formulate this as a regression task of forecasting future intermediate representation $\mathbf{\hat{F}}_{t +\Delta}$
of the proposed two-stream network based on its current intermediate representation $\mathbf{\hat{F}}_{t}$.
The main idea is that the intermediate representation of our proposed network abstracts spatial and motion information of hands and objects, and that we can learn a convolutional network modeling how such representation changes over time. 
Importantly, once we obtain the future intermediate representation $\mathbf{\hat{F}}_{t +\Delta}$,
we pass the predicted future representation to the decoder part of the auto-encoder as well as the remaining part of the SSD backbone network to explicitly forecast future hand/object bounding boxes. By feeding the regressed future intermediate representation, the SSD network will give the predicted coordinates of objects as if it has ``seen'' the future scene. 

Let $r$ denote our future regression network to predict
the future intermediate scene representation $\mathbf{\hat{F}}_{t +\Delta}$
given a current scene representation $\mathbf{\hat{F}}_{t}$.
\begin{equation} \label{eq:2}
\mathbf{\hat{F}}_{t +\Delta} = r_{w}(\mathbf{\hat{F}}_{t}).
\end{equation}
The regression network consist of nine convolutional layers, each having 256 channels of 5 $\times$ 5 filters except the last two layers.
We use dilated convolution with 1024 filters to cover a large receptive field of 13 $\times$ 13 for the 8th layer, and 256 1 $\times$ 1 filters are used for the last layer.

A desirable property of this formulation is that it allows training of the weights ($w$) of the regression network with unlabeled videos using the reconstruction loss as shown below:
\begin{align} \label{eq:3}
w^* &= \operatorname*{arg\,min}_{w}\sum_{i,t}\| r_{w}(g(\mathbf{\hat{X_I}}_{t}^i, \mathbf{\hat{X_O}}_{t}^i)) - g(\mathbf{\hat{X_I}}_{t +\Delta}^i, \mathbf{\hat{X_O}}_{t +\Delta}^i) \|_{2}^2
\end{align}
where $\mathbf{\hat{X}}^i_{t}$ indicates the frame or flow image at time $t$ from video $i$.
Once we get the future scene representation $\mathbf{\hat{F}}_{t +\Delta}$,
it is fed to $h$ to forecast hand/object locations corresponding to the future frame:
\begin{equation} \label{eq:4}
\mathbf{\hat{Y}}_{t+\Delta} = h(\mathbf{\hat{F}}_{t+\Delta}).
\vspace{-3pt}
\end{equation}

Figure \ref{fig:our_network} shows data flow of our proposed approach during the inference (i.e., testing) phase.
Given a video frame $\mathbf{\hat{X_I}}_{t}$ and its corresponding optical flow image $\mathbf{\hat{X_O}}_{t}$ at time $t$,
(1) we first extract the intermediate representation ($g$), and (2) give it to the future regression network ($r$) to obtain future scene representation $\mathbf{\hat{F}}_{t +\Delta}$.
Finally, (3) we predict future location of hands/objects $\mathbf{\hat{Y}}_{t+\Delta}$ by providing the predicted future scene representation to the remaining part of the proposed two-stream CNN ($h$) at time $t$.
\begin{equation} \label{eq:5}
\mathbf{\hat{Y}}_{t+\Delta} = h(\mathbf{\hat{F}}_{t+\Delta}) = h(r(\mathbf{\hat{F}}_{t}))  = h(r(g(\mathbf{\hat{X_I}}_{t}, \mathbf{\hat{X_O}}_{t}))).
\end{equation}

In addition to the above basic formulation, our proposed approach is extended to use previous $K$ frames to obtain $\mathbf{\hat{F}}_{t+\Delta}$ as illustrated in Figure \ref{fig:our_network}.
\begin{equation}
\mathbf{\hat{Y}}_{t+\Delta}
= h(r([g(\mathbf{\hat{X_I}}_{t}, \mathbf{\hat{X_O}}_{t}), ..., g(\mathbf{\hat{X_I}}_{t-(K-1)},\mathbf{\hat{X_O}}_{t-(K-1)} ])).
\end{equation}

Our future representation regression network allows predicting future objects
while considering the implicit scene/object/motion context in the scene.
The intermediate representation $\mathbf{\hat{F}}_{t}$ abstracts spatial/motion information in the current scene,
and our fully convolutional future regressor takes advantage of it for the forecast.

%% file: experiments.tex
\section{Experiments}

We conducted three sets of experiments to confirm the forecast ability of our approach using the fully convolutional two-stream regression architecture. In the first experiment, we use a first-person video dataset to predict future human hand locations. In the second and third experiments, we use one public egocentric video dataset and one street scene dataset with object annotations to evaluate our methods of predicting future object locations and presences.



\subsection{Dataset} \label{sec:dataset}


\vspace{-3pt}
\paragraph{\textbf{Human Interaction Videos:}}
We collected an in-house dataset which contains 47 first-person videos of human-human collaboration scenarios with a wearable camera. The dataset contains videos with two types of collaborative scenarios:
(1) a person wearing the camera cleaning up objects on a table
as another person approaches the table while holding a large box (i.e., making a room to put the box), and 
(2) the camera wearer pushes a trivet on a table as another person with a cooking pan approaches. The duration of each video clip is between 4 and 10 seconds. In our future location forecast task, we fix the SSD backbone including the auto-encoder part, and train the regressor part based on Section \ref{subsec:regression}.


\vspace{-3pt}
\paragraph{\textbf{Activities of Daily Living (ADL):}} 
This first-person video dataset \cite{pirsiavash2012detecting} contains 20 videos of 18 daily activities, such as making tea and doing laundry. 
This is a challenging dataset since frames display a significant amount of motion blur caused by the camera wearer's movement. This dataset also suffers from noisy annotations. Object bounding boxes were provided as ground truth annotations. Although there are 43 types of objects in the dataset, we trained our model (and the baselines) for the 15 most common categories, following the setting used in \cite{vondrick2015anticipating}. We split the ADL dataset into four sets, using three sets for the training and the remaining set for the testing, following the setting.

\vspace{-3pt}
\paragraph{\textbf{Cityscapes:}} 
This dataset~\cite{Cordts2015Cvprw,Cordts2016Cityscapes} is a benchmark dataset for semantic segmentation of objects in urban street scenes, which has 30 different classes of objects/scenes (e.g., person, car, sidewalk, ...) with both pixel-level and instance level annotations. For each video, the 20th image of every 30 frame video snippet has been annotated. That is, the time interval between two adjacent frames is 1.8 seconds.
We split 27 videos with fine-grained segmentation masks into 17/5/5 as our training/validation/test sets following the standard setting. We chose seven `object' classes with sufficient number of training instances in the dataset: bike, traffic sign, traffic light, rider, bus, car, and person. We evaluated forecasting future locations and presences of these objects as they are major moving objects in the scenes and also have enough numbers of occurrences in both train and testing sets (i.e., more than 200 samples). The original annotations are in forms of polygons as instance-level segmentation masks. We convert those polygons to bounding boxes to make our setting similar to the ADL forecast setting described above. Forecasting objects in the next annotated frame (1.8s later) and the next third annotated frame (5.4s later) was evaluated.
\vspace{-5pt}

\begin{figure*}[t]
 \centering
    \includegraphics[width=0.99\textwidth]{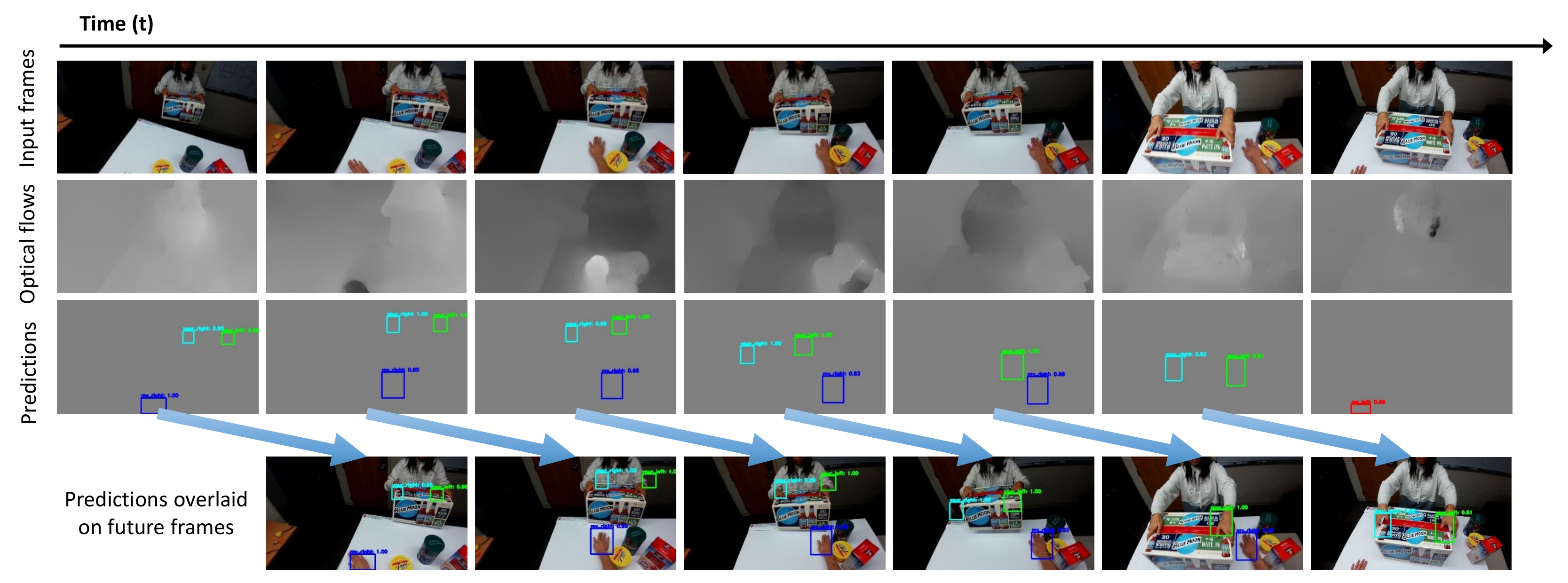}
    \vspace{0pt}
  \caption{Examples of hand location forecast.
          The first row shows the input frames and the second row shows the optical flows.
          The third row shows our future hand forecast results and we overlaid our predictions on `future' frames in the last row.
          Red boxes correspond to the predicted `my left hand' locations, blue boxes correspond to `my right hand', green boxes correspond to the opponent's left hand, and the cyan boxes correspond to the opponent's right hand. In the first frame, the model forecasts that the right hand `will appear', before it actually sees the hand.}
 \label{fig:hand_example}
\end{figure*}

\subsection{Baselines} \label{sec:baselines}
\vspace{-3pt}
In order to confirm the benefits of our proposed approach quantitatively, we created multiple baselines. 

\textbf{(i) SSD trained with future annotations} is the original SSD model \cite{ssd} for object bounding box estimation, which was trained to forecast future hands/objects. Instead of providing current-frame object bound boxes as ground truths in the training step, we provided `future' ground truth hand/object locations. This enables the model to directly regress future object boxes given the current frame. We also implemented a \textbf{ (ii) two-stream version of SSD}, making the SSD architecture to also consider optical flows.

\textbf{(iii) Hands only} is the baseline only using estimated hand locations in the current frame to predict their future locations. The idea is to confirm whether the detection of hand locations is sufficient to infer their future locations. A set of fully connected layers were used for the future location estimation, taking the current frame hand locations as its input representation.

In addition, we implemented simpler versions of our approach, \textbf{(iv) one-stream networks}, which use the same CNN architecture as our proposed approach except that it only has the spatial stream (taking RGB input) without the temporal stream (taking optical flow input). We constructed this baseline to confirm how much the temporal-stream of our network helps predicting future hand/object locations.
We also compare ours against \cite{vondrick2015anticipating} for future object presence forecast.
Finally, we are comparing our approach with the \textbf{(v) Vondrick et al.'s method} \cite{vondrick2015anticipating} designed for the object presence forecast. This was done by evaluating our approach on the ADL dataset with the same experimental setup as \cite{vondrick2015anticipating}.



\subsection{Training}

The training of our models was done in two stages. We first finetune the modified SSD network including the auto-encoder part (for scene representation) based on ground-truth object locations. Next, we train the future regressor (i.e., $r$) based on intermediate representations extracting from any \textit{current and future} frame pair in training videos, with the L2 loss (between $r(\mathbf{\hat{F}}_t)$ and $\mathbf{\hat{F}}_{t +\Delta}$) for measuring reconstruction errors. The second step is unsupervised learning, without requiring additional annotations. We found it more stable to separate these two stages than end-to-end training the entire model, since the 2nd stage can benefit more from unlabeled videos.

\subsection{Evaluation}

\paragraph{Hand location forecast:}
We first evaluated the performance of our approach to predict future hand locations using our unlabeled human interaction dataset. This is a less noisier dataset than the ADL dataset. Here, we use hand detection results (from the original SSD model trained on the EgoHands dataset \cite{Bambach_2015_ICCV}) as the ground truth hand labels for the evaluation, since the interaction videos do not have any human annotations. We randomly split the dataset into the training set and the test set; we used 32 videos for the training and the remaining 15 videos for the testing.\
We used the precision and recall as our evaluation measure. Whether the forecasted bounding boxes are true positives or not was decided based on the intersection over union (IoU) ratio between areas of each predicted box and the (future) ground truth box. The IoU threshold was 0.5.

\begin{table}[t]
\centering
\caption{Future hand \textit{location} forecasts measured with Human Interaction dataset.}
\vspace{-6pt}
{\scriptsize{\textsf{
\begin{adjustbox}{max width=0.99\linewidth}
\begin{tabular}{ l | l l l }
  \toprule
  \multirow{2}{*}{Method} & \multicolumn{3}{c}{Evaluation }\\
  \cmidrule{2-4}
   & Precision & Recall & F-measure \\
  \hline
   Hands only & 4.78 $\pm$ 3.70 & 5.06 $\pm$ 4.06 & 4.87 $\pm$ 3.81 \\
   SSD w/ future Annot. & 27.53 $\pm$ 23.36 & 9.09 $\pm$ 8.96 & 13.23 $\pm$ 12.62 \\
  \hline
   Ours (one-stream): K=1 & 27.04 $\pm$ 16.50 & 21.71 $\pm$ 14.71 & 23.45 $\pm$ 14.99 \\
   Ours (one-stream): K=5 & 29.97 $\pm$ 15.37 & 23.89 $\pm$ 16.45 & 25.40 $\pm$ 15.51 \\
   Ours (one-stream): K=10 & 36.58 $\pm$ 16.91 & 28.78 $\pm$ 17.96 & 30.90 $\pm$ 17.02 \\
   \hline
   Ours (two-stream): K=1 & 37.21 $\pm$ 22.49 & 26.69 $\pm$ 14.28 & 30.21 $\pm$ 16.07 \\
   Ours (two-stream): K=5 & 37.41 $\pm$ 22.97 & 26.19 $\pm$ 14.93 & 30.06 $\pm$ 17.16 \\
   Ours (two-stream): K=10 & \textbf{42.89} $\pm$ 23.61 & \textbf{30.46} $\pm$ 13.08 & \textbf{34.18} $\pm$ 16.48 \\
  \bottomrule
\end{tabular}
\end{adjustbox}
}}}
\label{table:hand_locations}
\end{table}

Table~\ref{table:hand_locations} shows quantitative results of 1-second future hand prediction.
Since our network may use previous $K$ frames as an input for the future regression, we reported the performances of our approach with $K$=$1,5,10$ frames. 
We observe that our proposed approaches significantly outperform the original SSD trained with future hand locations. The one-stream model performed better than the SSD baseline, suggesting the effectiveness of our concept of future regression. Note that our one-stream K=1 takes the exactly same amount of input as the SSD baseline. Our two-stream models performed better than the one-stream models, indicating the temporal stream is helpful to predict future locations. Our proposed model with $K=10$ yields the best performance in terms of all three metrics, at about 34.18 score in F-measure. Figure~\ref{fig:hand_example} shows example hand forecast results.



\vspace{-10pt}
\paragraph{Object location forecast:}
We used the ADL dataset~\cite{pirsiavash2012detecting} to evaluate future object location forecast performances. Both 1-second and 5-second future bounding box locations are predicted, and the performances were measured in terms of mean average precision (mAP). The IoU ratio of 0.5 was used to determine whether a predicted bounding box is correct compared to the ground truth. Note that ADL dataset is a challenging dataset for future prediction, since the videos were taken from the first-person view displaying strong egocentric motion. Further, appearing objects are not evenly distributed across different videos. Many objects appear and disappear from the scene even within the 5 second window due to the camera ego-motion.



Table~\ref{table:future_loc_prediction} shows average precision (AP) of each object category. We show that our approach significantly outperforms the SSD baseline. While only taking advantage of the same amount of information (i.e., a single frame), our approach (one-stream K=1) achieved a superior performance.
By using additional temporal information, our approach (two-stream K=1,10) outperforms its one-stream version by $2\text{-}5\%$ in mAP. This indicates that motion information is helpful in predicting the right location of objects in future frames, especially in first-person videos with strong ego-motion.
Figure~\ref{fig:adl_examples} shows example object predictions in 1-second and 5-second future. Based on RGB and optical flow information in the frames, our approach is able to predict future objects even when they are not visible in the current scene.

\begin{table*}[t]
\centering
\caption{Future object \textit{location} forecast evaluation using the ADL dataset.}
\vspace{-6pt}
 \begin{adjustbox}{max width=\textwidth}
 \begin{tabular}{ c | l | *{14}{c} | l }
    & Method &     
                 dish & door & utensil & cup & oven &
                 person & soap & tap & tbrush & tpaste &
                 towel & trashc & tv & remote  & mAP \\ \hline

    \multirow{6}{*}{1 sec} & SSD with future annotation &
                 0 & 0.5 & 0 & 0 & 0 &
                 0.2 & 0 & 1.3 & 0 & 0 &
                 0 & 0 & 0 & 0 & 0.1 \\
    & SSD (two-stream) &
                 1.6 & 12.4 & 0 & 0.9 & 5.1 &
                 9.6 & 0 & 2.8 & 0 & 0 &
                 0 & 0 & 29.8 & 1.3 & 4.5 \\
    & Ours (one-stream K=1) &
                 0.4 & 15.0 & 1.2 & 2.6 & 13.8 &
                 43.4 & 4.4 & 19.0 & 0 & 0 &
                 0.3 & 0 & 16.0 & 18.8 & 9.6 \\
    & Ours (one-stream K=10) &
                 0.4 & 14.0 & 0 & 0.7 & 16.1 &
                 45.8 & 5.4 & 22.9 & 0 & 0 &
                 0.9 & 0 & 20.5 & 6.8 & 9.5 \\
    & Ours (two-stream K=1) &
                 7.3&19.6&1.9&1.8&37.2&26.2&11.6&33.8&0.0&1.4&1.5&0.8&11.0&11.9&11.9 \\
    & Ours (two-stream K=10) &
                 3.8&10.1&1.8&5.5&19.0&59.6&2.8&41.8&0.0&0.0&3.4&0.0&15.9&45.2&\textbf{14.9} \\
    \hline
    \multirow{6}{*}{5 sec} & SSD with future annotation &
                 0 & 0 & 0 & 0 & 0.2 &
                 0 & 0 & 0 & 0 & 0 &
                 0 & 0 & 3.0 & 0 & 0.2 \\
    & SSD (two-stream) &
                 2.0 & 11.7 & 0 & 3.2 & 10.2 &
                 0.5 & 3.2 & 0 & 0 & 0 &
                 0 & 0 & 20.0 & 0 & 3.7 \\
    & Ours (one-stream K=1) &
                 0.5 & 10.8 & 0 & 0.3 & 16.5 &
                 10.4 & 3.2 & 8.2 & 0 & 0 &
                 0.7 & 0 & 3.9 & 1.7 & 4.0 \\
    & Ours (one-stream K=10) &
                 0.2 & 10.7 & 0 & 0.2 & 0.7 &
                 35.7 & 1.3 & 5.6 & 0 & 0 &
                 0.5 & 0 & 3.8 & 1.2 & 4.7 \\
    & Ours (two-stream K=1) &
                 1.5&9.8&0.4&0.4&24.1&17.0&8.6&15.8&0.0&0.0&1.6&0.2&7.5&5.8&6.6 \\
    & Ours (two-stream K=10) & 
                 0.7&4.7&0.0&5.0&9.7&35.6&0.7&10.5&0.0&0.0&1.4&0.0&15.0&24.8&\textbf{7.7} \\
    \hline
 \end{tabular}
 \end{adjustbox}
\label{table:future_loc_prediction}
\end{table*}

\begin{table*}[t]
 \caption{Future object \textit{presence} forecast (5sec) evaluation using the ADL dataset.}
\begin{adjustbox}{max width=\textwidth}
\begin{tabular}{ c | l | *{14}{c} | l }
    & Method &     
                dish & door & utensil & cup & oven &
                person & soap & tap & tbrush & tpaste &
                towel & trashc & tv & remote  & mAP \\ \hline

    \multirow{7}{*}{5 sec} & Vondrick \cite{vondrick2015anticipating} &
                 4.1 & 22.2 & 5.7 & 16.4 & 17.5 &
                 8.4 & 19.5 & 20.6 & 9.2 & 5.3 &
                 5.6 & 4.2 & 8.0 & 2.6 & 10.7 \\
    & SSD with future annotation &
                 18.9 & 17.6 & 0 & 28.1 & 7.1 &
                 23.0 & 0 & 37.7 & 0 & 0 &
                 0 & 0 & 20.4 & 0 & 10.9 \\       
    & SSD (two-stream) &
                 13.5 & 22.4 & 0 & 15.2 & 4.1 &
                 14.3 & 39.8 & 21.4 & 0 & 0 &
                 0 & 0.4 & 48.4 & 0 & 12.8 \\
    & Ours (one-stream K=1) &
                 34.4&37.0&18.9&19.2&24.3&75.1&70.0&55.0&23.8&6.7&16.6&2.1&57.5&61.7&35.9 \\
    & Ours (one-stream K=10) &
                 35.1&42.4&22.2&29.9&37.9&69.9&68.0&67.6&21.7&47.7&17.7&5.2&30.5&36.4&38.0 \\
    & Ours (two-stream K=1) &
                 38.2&44.1&23.8&29.1&37.2&73.1&67.1&60.6&12.2&38.0&13.7&4.4&37.2&58.5&38.4 \\             
    & Ours (two-stream K=10) &
                 35.7&44.0&24.2&29.3&39.6&75.7&68.9&63.2&20.4&47.2&18.2&4.6&40.4&60.3&\textbf{40.8} \\
    \hline
 \end{tabular}
 \end{adjustbox}
\label{table:future_presence_prediction_5s}
\end{table*}


\vspace{-10pt}
\paragraph{Object presence forecast:} 

In this experiment, we used the ADL dataset to evaluate our approach in forecasting `presence' of objects in future frames. 
Specifically, we decide whether the objects will \emph{exist} (in the future frame) or not, regardless their locations.
Similar to our object location forecast experiment, we obtained PR-curves and calculated AP of each object category. We trained our model to predict presence of objects in 5-second-future frames. This experiment makes it possible to directly compare our approach with the results of \cite{vondrick2015anticipating}'s AlexNet based architecture, following the same standard setting used in their experiments.

Table \ref{table:future_presence_prediction_5s} compares different versions of our proposed approach with the baselines. We observe that that our approaches significantly outperform the results reported in \cite{vondrick2015anticipating} while following the same setting. Our two-stream K=10 version obtained the mean AP of 40.8\%, which is higher than the previous state-of-the-art by the margin of 30\%. In addition, our one-stream K=1 version that only uses one single RGB frame as an input obtained higher accuracy than the SSD baseline and \cite{vondrick2015anticipating} while using the same input. Their performances were 35.9 vs. 10.9 vs. 10.7. We also confirmed that our two-stream K=1 version performs better than the two-stream version of SSD.





Table \ref{table:future_loc_prediction_city} shows additional experimental results on Cityscapes dataset, measured for both the location and presence forecast tasks. In this dataset, frames are annotated with 1.8 second gaps; the annotation frame rate is low. We thus only trained K=1 versions of our methods in this experiment. We compare our methods (both the one-stream version and the two-stream version) with the SSD two-stream baseline. We observe that our future regression models perform superior to the baseline two-stream SSD, by benefiting from their explicit future representation regression capability. This is consistent in both 1.8s and 5.4s forecast tasks, and in both the future location and presence prediction tasks.

\begin{table*}[t]
\centering
 \caption{Future object forecast evaluation using the Cityscapes dataset. Per-class AP varies a lot depending on what object class the CNN model decides to fire more frequently (i.e., learned prior/bias), but the mean AP results show more consistent trend.}
 \vspace{-6pt}
 \begin{adjustbox}{max width=\textwidth}
 \begin{tabular}{ c | c | *{9}{c} | c }
 \multicolumn{9}{c}{(a): Future object location}\\

    & Method &     
                 bike & sign & light & rider & bus & car & person & mAP \\ \hline

    \multirow{3}{*}{1.8 sec} & SSD (two-stream) &
                 3.6&2.3&0.0&0.2&0.2&14.8&12.8&4.9 \\
    & Ours (one-stream K=1) &
                 4.5&3.5&0.6&1.9&0.0&13.0&15.4&5.6 \\
    & Ours (two-stream K=1) &
                 7.8&2.6&0.3&7.2&0.0&11.6&15.1&6.4 \\
    \hline
    \multirow{3}{*}{5.4 sec} & SSD (two-stream) &
                2.3&1.4&0.1&0.3&0.3&15.2&7.4&3.9 \\
    & Ours (one-stream K=1) &
                 3.2&2.1&0.1&0.5&0.4&15.4&14.2&5.1 \\
    & Ours (two-stream K=1) &
                 7.2&2.3&0.9&5.1&1.0&11.3&10.0&5.4 \\
    \hline
 \end{tabular}
 \end{adjustbox}
 
 \vspace{6pt}

  \begin{adjustbox}{max width=\textwidth}
 \begin{tabular}{ c | c | *{9}{c} | c }
 \multicolumn{9}{c}{(b): Future object presence}\\

    & Method &     
                 bike & sign & light & rider & bus & car & person & mAP \\ \hline

    \multirow{3}{*}{1.8 sec} & SSD (two-stream) &
                  77.3&95.9&39.1&28.8&15.2&22.9&83.9&51.9 \\
    & Ours (one-stream K=1) &
                  74.7&94.2&55.4&45.6&7.8&20.6&88.0&55.2\\
    & Ours (two-stream K=1) &
                  75.6&92.1&54.0&64.8&6.8&14.9&84.0&56.1 \\
    \hline
    \multirow{3}{*}{5.4 sec} & SSD (two-stream) &
                  79.1&95.6&26.8&24.2&9.7&22.9&76.3&47.8 \\
    & Ours (one-stream K=1) &
                  73.8&93.4&45.5&60.3&9.6&15.0&77.8&53.6 \\
    & Ours (two-stream K=1) &
                  77.7&91.6&34.4&32.7&13.5&31.4&75.5&51.0 \\
    \hline
 \end{tabular}
 \end{adjustbox}
\label{table:future_loc_prediction_city}
\end{table*}

\begin{figure*}[b]
 \centering
    \includegraphics[width=0.9\textwidth]{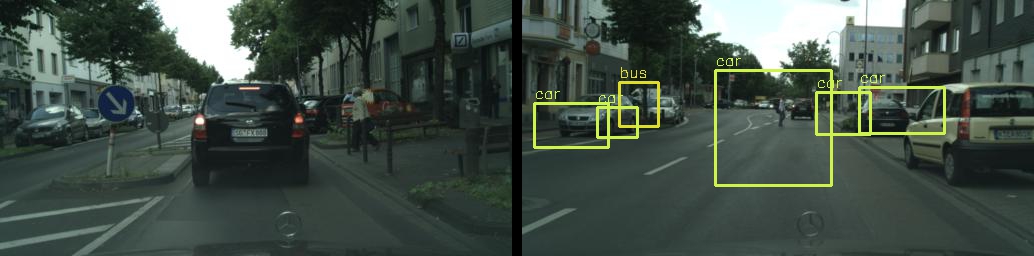}
    \includegraphics[width=0.9\textwidth]{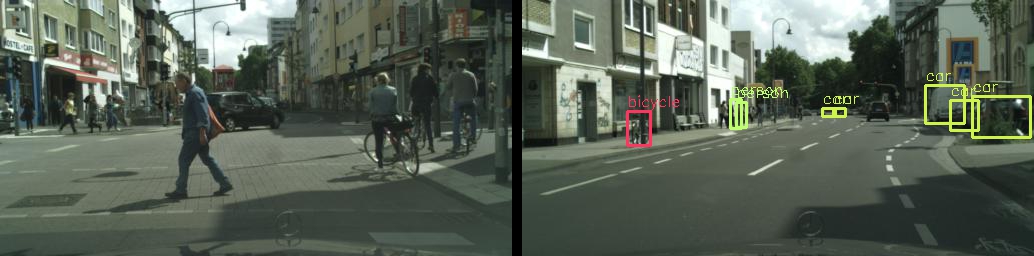}
    \vspace{0pt}
  \caption{Examples of street scene object forecast. Left column shows current frames, while right column shows frames after 1.8 seconds.}
 \label{fig:city_example}
\end{figure*}

%% file: conclusion.tex
\section{Conclusion}

We presented a new approach to explicitly \emph{forecast} human hands and objects using a fully convolutional future representation regression network. The key idea was to forecast scene configurations of the near future by predicting (i.e., regressing) intermediate CNN representations of the future scene. We presented a new two-stream model to represent scene information of the given image frame, and experimentally confirmed that we can learn a function (i.e., a network) to model how such intermediate scene representation changes over time. The experimental results confirmed that our object forecast approach significantly outperforms the previous work on the public ADL dataset.


%% file: eccvw18.bbl
\begin{thebibliography}{10}
\providecommand{\url}[1]{\texttt{#1}}
\providecommand{\urlprefix}{URL }
\providecommand{\doi}[1]{https://doi.org/#1}

\bibitem{Bambach_2015_ICCV}
Bambach, S., Lee, S., Crandall, D.J., Yu, C.: Lending a hand: Detecting hands
  and recognizing activities in complex egocentric interactions. In: IEEE
  International Conference on Computer Vision (ICCV) (December 2015)

\bibitem{Cordts2016Cityscapes}
Cordts, M., Omran, M., Ramos, S., Rehfeld, T., Enzweiler, M., Benenson, R.,
  Franke, U., Roth, S., Schiele, B.: The cityscapes dataset for semantic urban
  scene understanding. In: Proc. of the IEEE Conference on Computer Vision and
  Pattern Recognition (CVPR) (2016)

\bibitem{Cordts2015Cvprw}
Cordts, M., Omran, M., Ramos, S., Scharw{\"a}chter, T., Enzweiler, M.,
  Benenson, R., Franke, U., Roth, S., Schiele, B.: The cityscapes dataset. In:
  CVPR Workshop on The Future of Datasets in Vision (2015)

\bibitem{finn2016unsupervised}
Finn, C., Goodfellow, I., Levine, S.: Unsupervised learning for physical
  interaction through video prediction. In: Advances in Neural Information
  Processing Systems (NIPS). pp. 64--72 (2016)

\bibitem{hoai2012max}
Hoai, M., De~la Torre, F.: Max-margin early event detectors. In: IEEE
  Conference on Computer Vision and Pattern Recognition (CVPR) (2012)

\bibitem{kitani12}
Kitani, K.M., Ziebart, B.D., Bagnell, J.A., Hebert, M.: Activity forecasting.
  In: European Conference on Computer Vision (ECCV) (2012)

\bibitem{ssd}
Liu, W., Anguelov, D., Erhan, D., Szegedy, C., Reed, S., Fu, C., Berg, A.:
  {SSD}: Single shot multibox detector. In: European Conference on Computer
  Vision (ECCV) (2016)

\bibitem{lotter2016deep}
Lotter, W., Kreiman, G., Cox, D.: Deep predictive coding networks for video
  prediction and unsupervised learning. arXiv preprint arXiv:1605.08104  (2016)

\bibitem{luo2017unsupervised}
Luo, Z., Peng, B., Huang, D.A., Alahi, A., Fei-Fei, L.: Unsupervised learning
  of long-term motion dynamics for videos. arXiv preprint arXiv:1701.01821
  (2017)

\bibitem{ma2016trajectories}
Ma, W., Huang, D., Lee, N., Kitani, K.M.: A game-theoretic approach to
  multi-pedestrian activity forecasting. arXiv preprint arXiv:1604.01431
  (2016)

\bibitem{park2016future}
Park, H.S., Hwang, J.J., Niu, Y., Shi, J.: Egocentric future localization. In:
  IEEE Conference on Computer Vision and Pattern Recognition (CVPR) (2016)

\bibitem{pirsiavash2012detecting}
Pirsiavash, H., Ramanan, D.: Detecting activities of daily living in
  first-person camera views. In: IEEE Conference on Computer Vision and Pattern
  Recognition (CVPR). pp. 2847--2854. IEEE (2012)

\bibitem{rhinehart2017iccv}
Rhinehart, N., Kitani, K.M.: First-person activity forecasting with online
  inverse reinforcement learning. In: IEEE International Conference on Computer
  Vision (ICCV) (2017)

\bibitem{ryoo11}
Ryoo, M.S.: Human activity prediction: Early recognition of ongoing activities
  from streaming videos. In: IEEE International Conference on Computer Vision
  (ICCV) (2011)

\bibitem{simonyan2014two}
Simonyan, K., Zisserman, A.: Two-stream convolutional networks for action
  recognition in videos. In: Advances in Neural Information Processing Systems
  (NIPS). pp. 568--576 (2014)

\bibitem{vgg}
Simonyan, K., Zisserman, A.: Very deep convolutional networks for large-scale
  image recognition. In: ICLR (2015)

\bibitem{vondrick2015anticipating}
Vondrick, C., Pirsiavash, H., Torralba, A.: Anticipating visual representations
  with unlabeled video. In: IEEE Conference on Computer Vision and Pattern
  Recognition (CVPR) (2016)

\bibitem{walker2016uncertain}
Walker, J., Doersch, C., Gupta, A., Hebert, M.: An uncertain future:
  Forecasting from static images using variational autoencoders. In: European
  Conference on Computer Vision (ECCV) (2016)

\bibitem{walker2014patch}
Walker, J., Gupta, A., Hebert, M.: Patch to the future: Unsupervised visual
  prediction. In: IEEE Conference on Computer Vision and Pattern Recognition
  (CVPR). pp. 3302--3309 (2014)

\bibitem{visualdynamics16}
Xue, T., Wu, J., Bouman, K.L., Freeman, W.T.: Visual dynamics: Probabilistic
  future frame synthesis via cross convolutional networks. In: Advances in
  Neural Information Processing Systems (NIPS) (2016)

\bibitem{yagi2017future}
Yagi, T., Mangalam, K., Yonetani, R., Sato, Y.: Future person localization in
  first-person videos. In: IEEE Conference on Computer Vision and Pattern
  Recognition (CVPR) (2018)

\end{thebibliography}
